\documentclass[letterpaper, 10 pt, conference]{ieeeconf}  
\usepackage[T1, OT1]{fontenc}
\usepackage{booktabs}
\DeclareTextSymbolDefault{\DH}{T1}

\IEEEoverridecommandlockouts                              

\usepackage{cite}
 \usepackage{multirow}
\usepackage{algorithmic}
\usepackage{subfig}
\usepackage[ruled]{algorithm}

\usepackage{mathtools}
\usepackage{amssymb}
\usepackage{float}

\usepackage{graphicx}
\usepackage[export]{adjustbox}

\title{\LARGE \bf Deep Semantic Segmentation at the Edge for Autonomous Navigation in Vineyard Rows
}

\author{Diego Aghi$^{1}$, Simone Cerrato$^{2}$, Vittorio Mazzia$^{3, *}$ and Marcello Chiaberge$^{4}$
\thanks{*Corresponding author}
\thanks{$^{1}$Diego Aghi is with the Department of Electronics and Telecommunications,
        Politecnico di Torino, Turin, Italy 10124
        {\tt\small diego.aghi@polito.it}}%
\thanks{$^{2}$Simone Cerrato is with the Department of Electronics and Telecommunications,
        Politecnico di Torino, Turin, Italy 10124
        {\tt\small simone.cerrato@polito.it}}%
\thanks{$^{3}$Vittorio Mazzia is with the Department of Electronics and Telecommunications,
        Politecnico di Torino, Turin, Italy 10124
        {\tt\small vittorio.mazzia@polito.it}}%
\thanks{$^{4}$Marcello Chiaberge is with the Department of Electronics and Telecommunications,
        Politecnico di Torino, Turin, Italy 10124
        {\tt\small marcello.chiaberge@polito.it}}%
}

\begin{document}

\maketitle
\thispagestyle{empty}
\pagestyle{empty}

\begin{abstract}
Precision agriculture is a fast-growing field that aims at introducing affordable and effective automation into agricultural processes. Nowadays, algorithmic solutions for navigation in vineyards require expensive sensors and high computational workloads that preclude large-scale applicability of autonomous robotic platforms in real business case scenarios. From this perspective, our novel proposed control leverages the latest advancement in machine perception and edge AI techniques to achieve highly affordable and reliable navigation inside vineyard rows with low computational and power consumption. Indeed, using a custom-trained segmentation network and a low-range RGB-D camera, we are able to take advantage of the semantic information of the environment to produce smooth trajectories and stable control in different vineyards scenarios. Moreover, the segmentation maps generated by the control algorithm itself could be directly exploited as filters for a vegetative assessment of the crop status. Extensive experimentations and evaluations against real-world data and simulated environments demonstrated the effectiveness and intrinsic robustness of our methodology.
\end{abstract}


\section{INTRODUCTION}
Over the last years, the agriculture industries have been focusing their resources on the study and development of new technologies to increase productivity, cut costs, and ease farmers' jobs by reducing the need for humans for labor-intensive tasks. The wave of innovation brought by the advent of precision agriculture and digital farming has gradually introduced robotics and artificial intelligence into agricultural processes to improve product quality and management. In this context, self-driving systems for agricultural machineries have been a great breakthrough towards the accomplishment of the objectives mentioned above. Indeed, once endowed the proper equipment, these robotic vehicles can harvest \cite{harvesting}, spray\cite{spraygrape}, seed\cite{seeding} and irrigate\cite{irrigation} as well as collect data for inventory management\cite{monitoring}. 

Concerning the autonomous navigation in vineyards and orchards, some of the proposed solutions employ GPS devices along with three-dimensional LIDARs and Inertial Measurement Units (IMU) \cite{lidar_gps}, while others exploit only 2D LiDARs \cite{lidar_based_nav}\cite{lidar_based_nav2}. Nonetheless, they face difficulties to reach a large-scale implementation due to the elevated sensors costs and the unreliability in particular situations, which will be discussed later in this work. These issues made the research community move its focus on different kinds of technologies. Indeed, the state-of-the-art approaches for row crop scenarios with thick and high canopies are based on machine vision \cite{RADCLIFFE2018165}, and deep learning \cite{aghi2020autonomous}. 
However, in \cite{RADCLIFFE2018165} they use a multispectral camera for image acquisition which significantly raises hardware costs, while in \cite{aghi2020autonomous} the controller provided is not proportional, and it is limited to 3 basics commands.

In this article, we present a novel low-cost and at the edge motion control system for autonomous navigation in vineyard rows. It exploits semantic segmentation properties to provide a proportional controller that drives the robotic platform along the whole row without colliding with the vine plants. Moreover, it fuses RGB images and depth information of the observed scene to overcome illumination issues and provide a robust control in challenging situations without relying on expensive sensors and GPS signals. Finally, it is possible to directly take advantage of segmentation maps generated by the deep neural network for a vegetative real-time assessment of the crop.

All of our training and testing code and data\footnote{10.5281/zenodo.4601472} are open source and
publicly available\footnote{https://github.com/MrD1360/deep\_segmentation\_vineyards\_navigation}.

\subsection{Related Work}
Generally, autonomous systems designed to navigate along vineyards or orchards rows employ high-precision GPS receivers and enhancement accuracy techniques \cite{gps_only} or a combination of laser-based sensors and GPS devices\cite{lidar_gps2,hansen2011orchard}. However, in this particular environment, the canopies on the sides of the row reduce the GPS accuracy affecting its reliability\cite{GPS_accuracy,gpsunreliable}. Modern solutions mix multiple sensors such as GPS, inertial navigation systems (INS), wheel encoders, and LIDARs \cite{echord} to locate more precisely the mobile platform during the navigation. Nevertheless, the usage of many sensors leads to higher system costs. 

Regarding more affordable approaches, in \cite{zaman2019cost} they propose a cost-effective monocular visual odometry (VO) algorithm. Nonetheless, as highlighted by the authors, VO systems show poor performance on long distances due to the accumulating error, and they require an absolute reference integration to preserve the necessary accuracy for a safe autonomous navigation.

On the other hand, the ubiquity of deep learning in the latest technology advancements led to the development of a variety of intelligent systems for multiple applications in precision agriculture \cite{KAMILARIS201870}. Indeed, in literature can be found crop type classifiers\cite{cropclass1, cropclass2,mazzia2020improvement} and crop yield estimators\cite{yield_estimation,estimators2,mazzia2020uav} as well as disease detectors for plants \cite{desease_detector}, leaves\cite{leaf_desease} and fruits\cite{fruit_desease}.

\subsection{The Broader Project}
\begin{figure}
      \centering

    \includegraphics[width=1\linewidth]{"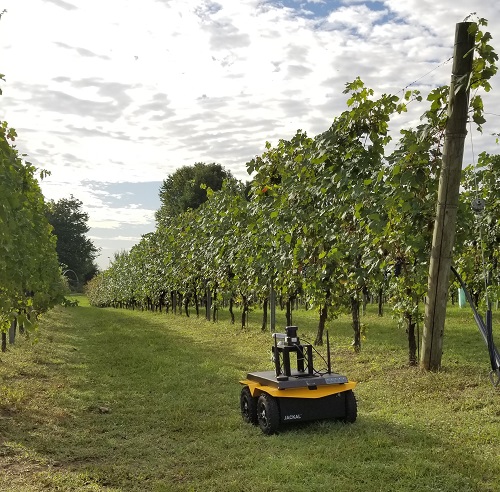"}
      \caption{The Jackal Unmanned Ground Vehicle in Grugliasco, Turin, North of Italy.}
      \label{jackal}
   \end{figure}

The proposed high-level controller for autonomous navigation in the vineyard rows is part of a broader project of our research group that aims at developing an affordable self-driving system for vineyard parcels.
Given a global path made of georeferenced waypoints \cite{deepway}, we exploit a dual local planner to overcome the lack of accuracy of the GPS-based localization filter inside the vine rows. Indeed, the signals provided by the satellites are very sensitive to obstacles encountered during the satellite-receiver path (e.g. lush vegetation of vineyards) and that makes the navigation inside vineyard rows a very challenging task \cite{GPS_accuracy,gpsunreliable,simo_thesis}.
More in detail, since outside the vineyard rows the GPS receiver (a Piksi Multi GNSS receiver by Swift Navigation) has a good view of the sky and better accuracy, a simple Dynamic Window Approach (DWA) is used to switch from one vine row to the next one, exploiting an Extended Kalman Filter (EKF) to fuse the data coming from an Inertial Measurement Unit (IMU) and a GPS receiver.
On the other hand, the proposed motion controller exploits semantic information of the vineyard environment in order to navigate throughout the vine rows. It performs periodic checks with respect to the provided global path made of GPS points, obtaining a broader estimation of its position. The overall solution guarantees to autonomously navigate throughout the whole vineyard without expensive sensors and in case of poor quality Global Navigation Satellite System signals due to lush vegetation and thick canopies.\\
Eventually, all the algorithms have been developed ROS-compatible, in order to make easier the communication among them and to be easily deployed on our developing platform; the Jackal Unmanned Ground Vehicle (UGV) by Clearpath Robotics\footnote{https://clearpathrobotics.com/}, shown in Fig. \ref{jackal}, that is fully supported by the Robot Operating System(ROS).

\section{METHODOLOGY}

\begin{figure*}
      \centering

    \includegraphics[width=1\linewidth]{"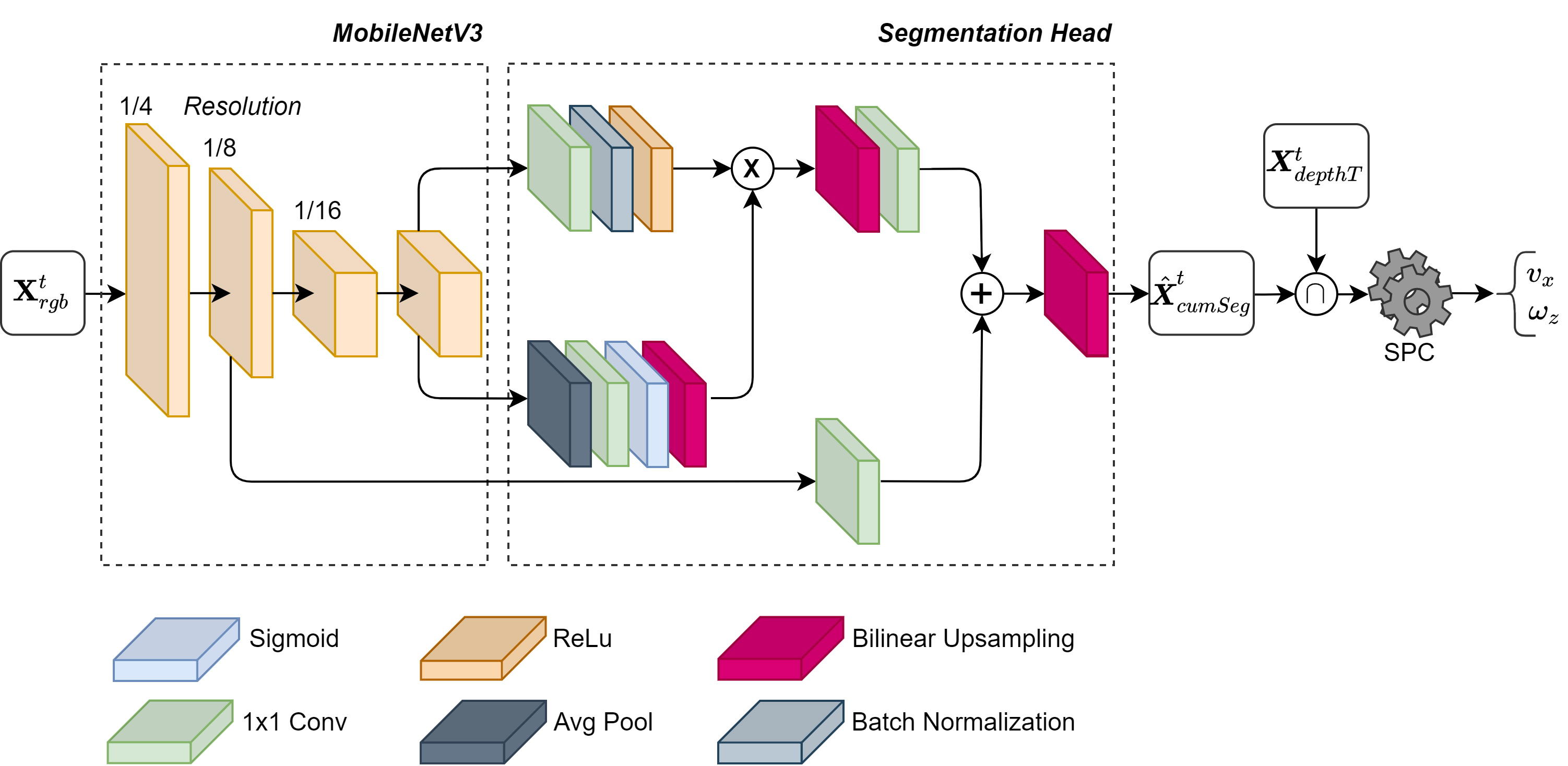"}
      \caption{The pipeline of our proposal. First, we feed the network with the RGB frame acquired by the camera to extract meaningful features. Next, the customized segmentation head provides the segmented frame combining features from different resolutions. Successively, we fuse S segmentation maps and intersect the resulting matrix with the thresholded depth map. Finally, the SPC takes as input the obtained binary map and it computes the control values for the autonomous vehicle.}
      \label{fig:architecture}
   \end{figure*}

We propose a control algorithm to perform real-time autonomous navigation along the vineyard rows that leverages the latest advancement on edge AI to obtain a continuous and reliable control with a low-range hardware setup. The devised system exploits the joint information of RGB and depth inputs, cutting costs of expensive sensors and overcoming issues experienced by solutions based on GPS devices.

The workflow of our proposal, as illustrated in Fig. \ref{fig:architecture}, is very straightforward; first, the RGB-D camera placed over the mobile platform acquires the depth map $\textbf{\textit{X}}_{depth}^{t}$ and the corresponding RGB frame $\textbf{X}_{rgb}^{t}$ at a certain time instant $t$, where $\textbf{\textit{X}}_{depth}^{t} \in \mathbb{R}^{h\times w}$ and $\textbf{
X}_{rgb}^{t} \in \mathbb{R}^{h\times w\times c}$ with $h$, $w$ and $c$ as high, width and channels, respectively. Then, we feed a custom deep neural network for semantic segmentation that produces a binary mask as shown in the following equation,
\begin{equation}
    \hat{\textbf{\textit{X}}}_{seg}^{t} = H_{seg}(\textbf{X}_{rgb}^{t}, \Theta)
    \label{network}
\end{equation}
where $\hat{\textbf{\textit{X}}}_{seg}^{t} \in  \mathbb{R}^{h\times w}$ is the segmentation map containing the semantic information of the input image $\textbf{X}_{rgb}^{t}$ and $\Theta$ are the model variables leaved to discriminative learning.
\begin{equation}
    \hat{\textbf{\textit{X}}}_{cumSeg}^{t} = \sum_{n=0}^{S} \hat{\textbf{\textit{X}}}_{seg}^{t-n}
    \label{sum_seg_maps}
\end{equation}
Successively, in order to obtain a more stable and coherent information, we select $S$ consecutive segmentation maps at times $\{t-S,...,t\}$ and we fuse them as shown in (\ref{sum_seg_maps}).
Then, we join information coming from the depth channel $\textbf{\textit{X}}_{depth}^{t}$ to dynamically reduce the line of sight of the processed scene. Indeed, that is a fundamental point to achieve an effective and reliable navigation inside curved vineyard rows. Moreover, the line of sight reduction is dynamically computed in order to suitably adjust the depth in different situations and discard as less information as possible from the segmented scene. That is achieved finding the maximum value in the depth matrix $max(\textbf{\textit{X}}_{depth}^{t})$ and generating a binary map $\textbf{\textit{X}}_{depthT}^{t}$, as follows:

\begin{equation}
\label{depth}
    {\textit{X}_{depthT_{\substack{i=0,...,h\\j=0,...,w}}}^{t}}(i,j)
    =\begin{cases} 0, & \mbox{if }(\textit{X}_{depth}^{t})_{i,j}\geq d_{depth} \\ 1, & \mbox{if }(\textit{X}_{depth}^{t})_{i,j}<d_{depth}
\end{cases}
\end{equation}

where $d_{depth} = l_{depth}*max(\textbf{\textit{X}}_{depth}^{t})$ and $l_{depth} \in \mathbb{R}$ is a scalar between 0 and 1. So, it is possible to limit the line of sight with an interception operation between the cumulative output of the network and the binary map generated,
\begin{equation}
    \textbf{\textit{X}}_{ctrl}^{t} = \sum_{n=0}^{S} \hat{\textbf{\textit{X}}}_{seg}^{t-n} \cap \textbf{\textit{X}}_{depthT}^{t}
\end{equation}
where $\textbf{\textit{X}}_{ctrl}^{t}\in  \mathbb{R}^{h\times w}$ represents the pre-processed input for the control algorithm. In the $\textbf{\textit{X}}_{ctrl}^{t}$ binary map $1$ means obstacles and $0$ free-space.

Finally, the actuation values for linear and angular velocity, $v_{x}$ and $\omega_{z}$, are calculated with a simple algorithm that extracts from the segmentation map a proportional control for the two variables.
\begin{equation}
    v_{x},\omega_{z} = \textrm{SPC}(\textbf{\textit{X}}_{ctrl}^{t})
    \label{spce_controll}
\end{equation}
In (\ref{spce_controll})$, \textrm{SPC}$ is the segmentation to proportional control algorithm that extracts from the pre-processed segmentation map, $\textbf{\textit{X}}_{ctrl}^{t}$, a continuous control for the platform. In addition, $v_x$ is the linear velocity along $x$ axis with respect to the robot's frame and $\omega_z$ is the angular velocity around $z$ axis with respect to the UGV's frame following the right-hand rule.

The overall solution can be easily integrated with a generic global path planner to accomplish an affordable and lightweight autonomous navigation along the whole vineyard parcel.

\subsection{Segmentation Network Architecture}
A segmentation network is usually composed of an initial feature extractor block and a final segmentation head that takes high-level representations and generates the corresponding segmentation map of the learned classes.
Over the years, several deep learning backbones for feature extraction have been proposed. Among all, we carefully select a model that guarantees high accuracy levels by also containing hardware costs and computational load. In particular, we use MobileNetV3\cite{mobnet3} as network backbone with a custom segmentation head for generating the segmentation map predictions $\hat{\textbf{\textit{X}}}_{seg}^{t}$ at a certain time instance $t$. Due to the efficient design of the overall network, the footprint on the RAM is greatly reduced as well as the computational workload required for inference. Overall, the selected backbone is composed of 15 blocks. Each of them presents a linear block and an inverted residual structure introduced with the second version of the network, MobileNetV2\cite{mobilenetv2}. Moreover, attention modules based on squeeze and excitation\cite{squeeze} are implemented in some of the residual layers and with different non-linearity depending on the specific layer. As in \cite{mobnet3}, our proposed segmentation head is an upgrade of the reduced version of the Atrous Spatial Pyramid Pooling module\cite{chen2014semantic,chen2017rethinking,mobilenetv2} (R-ASPP) called Lite R-ASPP (LR-ASPP). It includes two branches connected to different resolutions in order to capture information from multiple-level features. More specifically, one layer applies atrous convolution \cite{chen2017deeplab} to the 1/16 resolution to extract denser features, and the other one is used to add a skip connection \cite{skip} from the 1/4 resolution to work with more detailed information. 
Due to the fact that in \cite{mobnet3} they use a higher input dimension when performing semantic segmentation, we rescale the global average pooling layer setting the kernel size to 12$\times$12 with strides (4,5). Additionally, to have equal input and output dimensions, we add a bilinear upsampling of a factor of 8 at the end of the segmentation head.

\subsection{SPC Algorithm}
\begin{algorithm}[t]
	\caption{SPC algorithm}
	\label{alg:pseudocode}
	\begin{algorithmic}[1]
	    \REQUIRE{\textbf{\emph{X}$_{ctrl}^{t}$}: Pre-processed segmented image}
	    \ENSURE{\textbf{$v_{x}$},\textbf{$\omega_{z}$}: Continuous control commands}
		\STATE {noise\_reduction\_function()}
		\FOR {i=1,$\cdots$ w}
	    	\STATE{ $ \textbf{\textit{c}}\leftarrow$\textbf{sum}\_colums($\textbf{\textit{X}}_{ctrl}^{t}$)}
		\ENDFOR
		\STATE{removing\_small\_zero\_clusters($\textbf{\textit{c}}$)}
		\STATE{count\_zero\_clusters($\textbf{\textit{c}}$) }
		\STATE{\textbf{with} previous\_cluster\_center \textbf{as} pcc:}
		\IF {\textbf{not} anomaly\_detection()}
		    \STATE{compute\_cluster\_center()}
		    \STATE{$v_{x}$,$\omega_{z} \leftarrow$ control\_function()} 
		\ELSIF{ initial\_state }
		    \STATE{remove\_clusters\_from\_sides()}
		    \STATE{\textbf{max (}cluster,  \textbf{key}=len)}
		    \STATE{compute\_cluster\_center()}
		    \STATE{$v_{x}$,$\omega_{z} \leftarrow$ control\_function()} 
		
	    \ELSIF{pcc \textbf{is in} clusters  \OR
	    pcc \textbf{is near} clusters}
	        \STATE{select\_new\_cluster()}
	        \STATE{compute\_cluster\_center()}
		    \STATE{$v_{x}$,$\omega_{z} \leftarrow$ control\_function()} 
		\ENDIF
	\end{algorithmic}
\end{algorithm}

In Algorithm \ref{alg:pseudocode}, is described a schematic representation of the overall segmentation to proportional algorithm that generates a continuous control for the platform starting from a pre-processed cumulative segmentation output $\textbf{\textit{X}}_{ctrl}^{t}$. Firstly, we remove the noise due to grass on the terrain that could mislead our model when predicting by analyzing the network output. We sum the values over rows of $\textbf{\textit{X}}_{ctrl}^{t}$  obtaining a column 1D-array $\textbf{\textit{g}}_{noise}\in  \mathbb{R}^{h}$, then we select all the indices where $\textbf{\textit{g}}_{noise}<{th}_{noise}$, with ${th}_{noise}$ as threshold. Finally, we set ${\textbf{\textit{X}}_{ctrl}^{t}}_{(indices,:)}=0$.  Ideally, we would not have any ones in the top of the image and on the bottom, whilst the majority of them are supposed to be in the central belt.

After that, we sum the values over columns of the obtained matrix $\textbf{\textit{X}}_{ctrl}^{t}$, in order to generate the array $\textbf{\textit{c}}\in  \mathbb{R}^{w}$, that contains the amount of detected vineyards for each column. Therefore, every zero in $\textbf{\textit{c}}$ is a potential empty space where to route the mobile platform.
Next, we detect the clusters of zeros, which are the groups of consecutive zeros, in $\textbf{\textit{c}}$, and those with a length below a certain threshold are considered as unreliable, and therefore they are not taken into account when computing the control values.
Successively, we begin evaluating the different scenarios. First of all, we perform anomaly detection by counting the number of clusters. In case we have just one, we move forward with the control functions. In contrast, when dealing with more clusters, the algorithm tries to identify which is the right one by using information from the previous iterations. More specifically, it checks if the previously computed abscissa used in the control functions is contained or near one of the current detected clusters. In case it is the first algorithm execution, there is no previous command; therefore, supposing that the mobile platform is placed pointing at the center of the vine row, we remove the clusters laying on the sides of the matrix $\textbf{\textit{X}}_{ctrl}^{t}$.
After that, if there is still more than one cluster, we select the largest one. 
During the whole process, at any moment, if there are no valid clusters, the computed matrix $\textbf{\textit{X}}_{ctrl}^{t}$ is discarded and the next one is elaborated.

The identified cluster represents the obstacle-free space in which we can route the mobile platform to continue the navigation safely. The next step is to compute the linear and angular velocities to drive the mobile platform. To do so, we take the center of the selected cluster, that ideally corresponds to the center of the row in front of the UGV. After that, we estimate the velocities with two custom functions.
\begin{equation}
\label{eq1}
\omega_z =\begin{cases} - \omega_{z,max}  \cdot \left[ \frac{d^{2}}{(\frac{w}{2})^{2}} \right ], & \mbox{if }d\geq 0 \\ \omega_{z,max}  \cdot \left [ \frac{d^{2}}{(\frac{w}{2})^{2}}\right ], & \mbox{if }d<0
\end{cases} 
\end{equation}
 In (\ref{eq1}) is represented the control function for the angular velocity whilst for the linear velocity we use  (\ref{eq2}) as in \cite{aghi2020local}. 

\begin{equation}
\label{eq2}
v_{x} = v_{x,max} \cdot \left [ 1 - \left [\frac{d^{2}}{(\frac{w}{2})^{2}} \right ] \right ] 
\end{equation}
where $\omega_{z,max}$ and $v_{x,max}$ are two constants which defines the maximum angular and linear velocity of the mobile platform respectively, $w$ is the width of $\textbf{\textit{X}}_{ctrl}^{t}$ and \emph{d} is defined as:
    \begin{equation}
    \label{eq3}
    d = x_{c} - \frac{w}{2}
    \end{equation} 

with $x_{c}$ center coordinate of the selected cluster.
The final control values sent to the actuators are calculated using the exponential moving average (EMA), formalized in (\ref{eq4}), in order to prevent the mobile platform from sharp motion.

    \begin{equation}
    \label{eq4}
   EMA_t=EMA_{t-1} \cdot (1 - \alpha_{EMA}) + \begin{bmatrix} v_x \\ \omega_z  \end{bmatrix} \cdot \alpha_{EMA}
    \end{equation} 
where $t$ is the time step and $\alpha_{EMA}$ the multiplier for weighting the EMA.
\section{EXPERIMENTAL RESULTS AND DISCUSSION}
\subsection{Dataset Creation}
\label{Dataset_sect}

In order to create a dataset for training and testing the deep neural network, we carry out field surveys in two distinct agricultural areas in the North of Italy: Grugliasco near Turin in Piedmont region and Valle San Giorgio di Baone in the Province of Padua in the Veneto region. The data is collected at different times of the day, with diverse weather conditions, and they present a variety of terrain types and wine qualities.
To have different perspectives inside the vineyard rows, we acquire several videos with only three fixed orientations: one pointing the camera at the center of the vineyard row and the other two pointing to the left and right sides, respectively.
For training and testing, we select one frame every 25 in order to work with quite different scenarios.  
Finally, to generate the masks for the supervised training, we exploit the collected images by first manually annotating them using an open-source software\cite{dutta2019vgg}, and successively refining the obtained binary masks using a Gaussian mixture model\cite{gaussian} in order to train and evaluate the network with more accurate masks.
Before feeding the network, all the acquired images are resized to a fixed input dimension, 224x224, and then normalized with values from 0 to 1. 

Overall, our dataset consists of 1538 RGB images with their corresponding binary mask, of which 1038 from the vineyards in Veneto region and 500 from the other rural area.

\begin{figure}
\centering

\subfloat[]{
\includegraphics[scale=0.39]{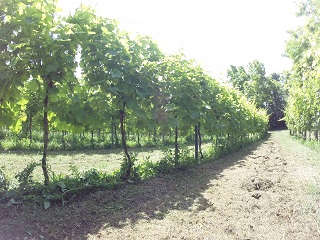}
\label{fig:fig21}
}
\subfloat[]{
\includegraphics[scale=0.39]{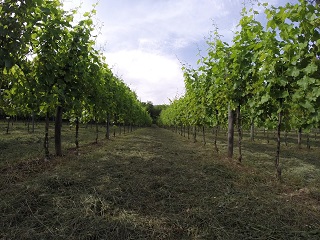}
\label{fig:fig22}
}
\subfloat[]{
\includegraphics[scale=0.39]{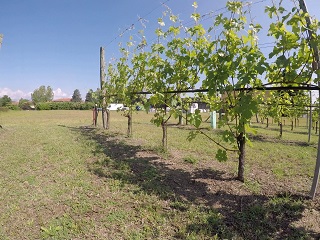}
\label{fig:fig23}
}\\
\subfloat[]{
\includegraphics[scale=0.39]{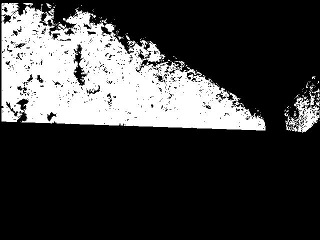}
\label{fig:fig24}
}
\subfloat[]{
\includegraphics[scale=0.39]{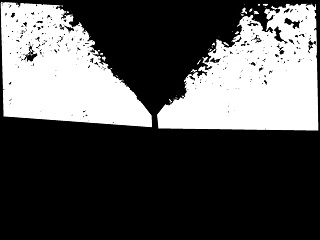}
\label{fig:fig25}
}
\subfloat[]{
\includegraphics[scale=0.39]{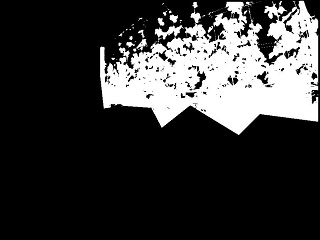}
\label{fig:fig26}
}

\caption{ In (\textbf{a}),(\textbf{b}) and (\textbf{c}) are shown three instances of the dataset, one for each video orientation. (\textbf{a}) is an example left samples, (\textbf{b}) center, and (\textbf{c}) right. (\textbf{d}),(\textbf{e}) and (\textbf{f}) are the corresponding binary masks for the supervised training.
Dataset samples have been collected with different weather conditions and at different times of the day. The resulting heterogeneous training set is aimed at giving generality and robustness to the model.}
\label{fig:datasetSample}
\end{figure}

\subsection{Network Training and Evaluation}
\begin{figure}[b]
\centering

\subfloat[]{
\includegraphics[scale=0.168]{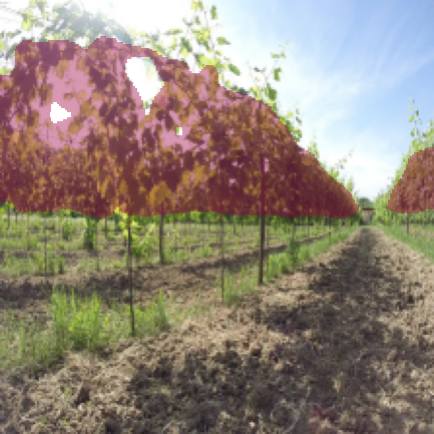}
\label{fig:fig61}
}
\subfloat[]{
\includegraphics[scale=0.168]{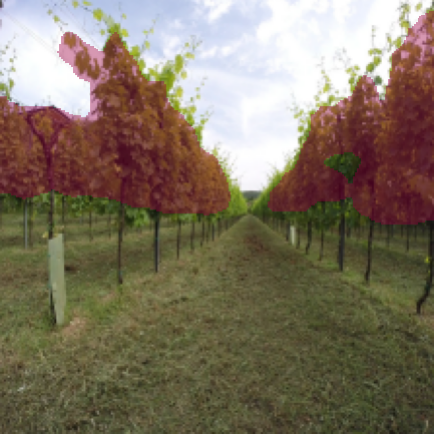}
\label{fig:fig62}
}
\subfloat[]{
\includegraphics[scale=0.168]{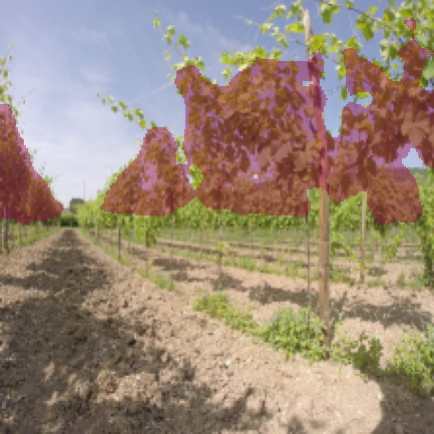}
\label{fig:fig63}
}\\
\subfloat[]{

\includegraphics[width=1.00in,height=1in,frame]{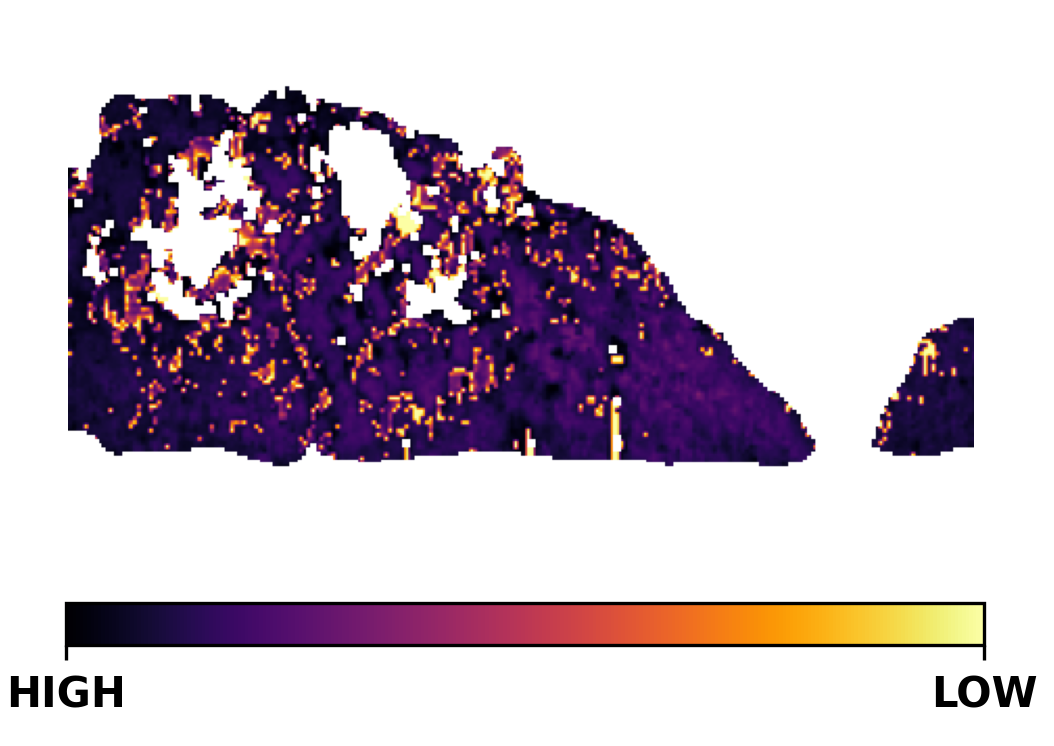}
\label{fig:fig64}
}
\subfloat[]{

\includegraphics[width=1.0in,height=1in,frame]{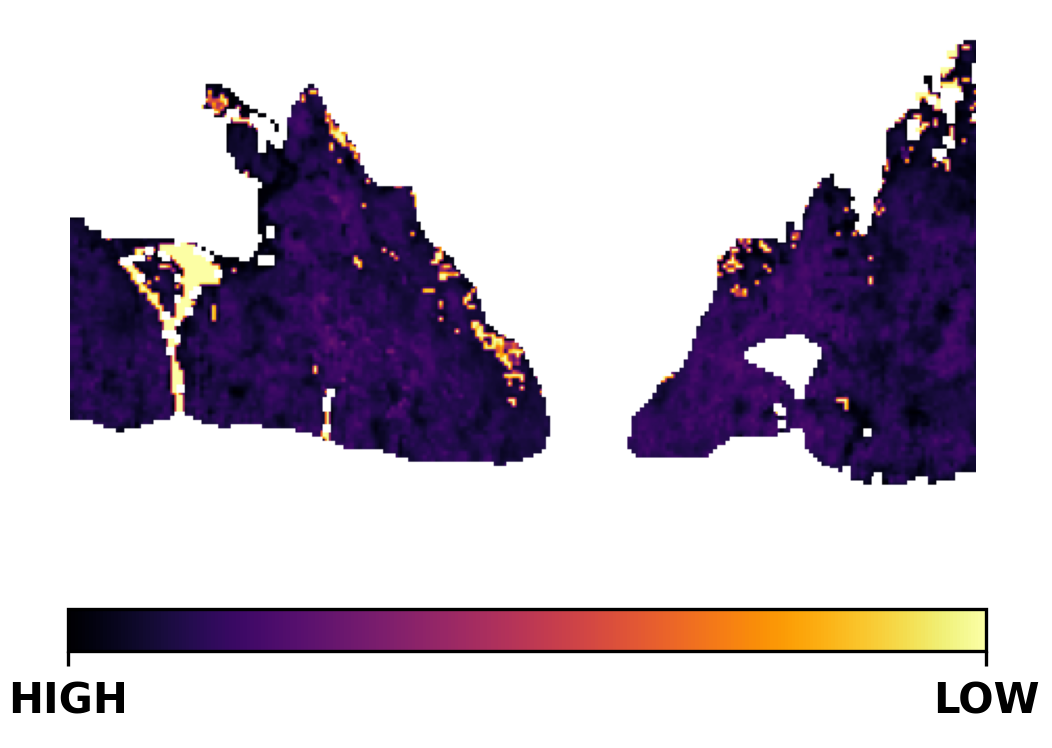}
\label{fig:fig65}

}
\subfloat[]{

\includegraphics[width=1.0in,height=1in,frame]{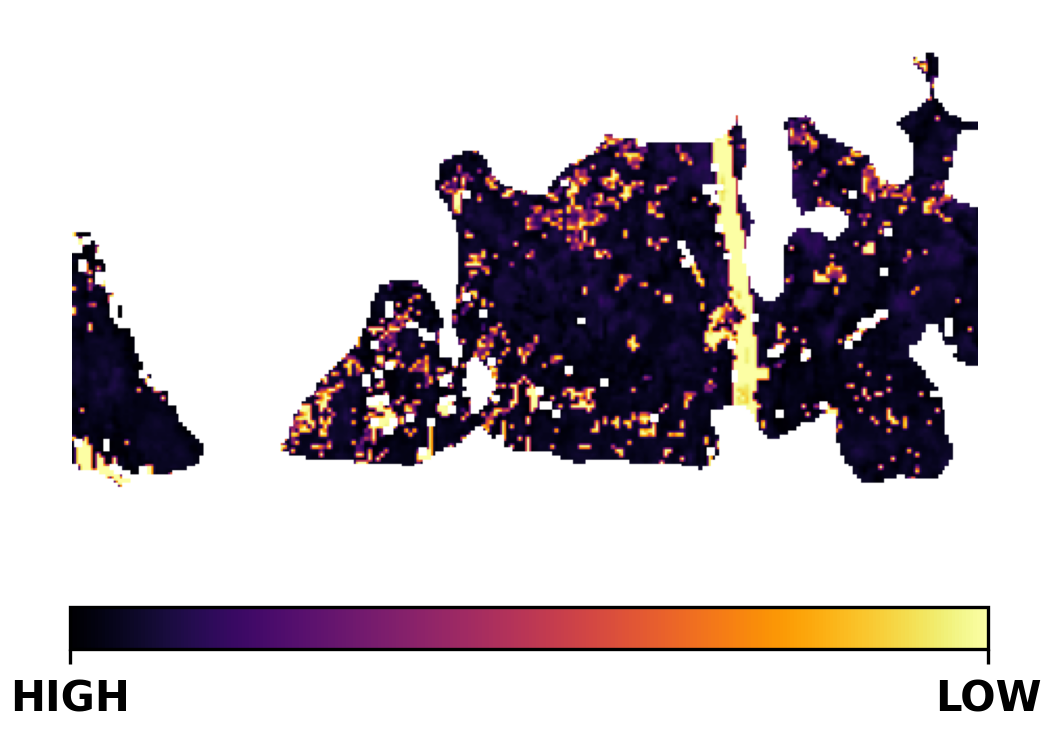}
\label{fig:fig66}
}

\caption{Qualitative assessment of the segmentation capabilities of the network. In (\textbf{a}), (\textbf{b}) and (\textbf{c}) are presented three samples taken from the test site. On the other hand, (\textbf{d}), (\textbf{e}) and (\textbf{f}) show a vegetative assessment that can be derived directly using generated masks and RGB channels \cite{padua2018multi}.} 
\label{fig:vegetation_index}
\end{figure}
For training and testing our segmentation network, we use the two surveyed rural areas. Indeed, a completely different area for testing allows to properly evaluate the generalization capabilities of the network not only in different weather conditions and times of the day but also in entirely different scenarios.

We train our model applying transfer learning \cite{tl} to the selected backbone. Indeed, rather than using randomly initialized weights, we exploit MobileNetV3 variables derived from an initial training phase on the 1k classes and 1.3M images of the ImageNet dataset\cite{russakovsky2015imagenet}. That largely improves the final robustness of the model and its final generalization capability with a reduced number of training samples. 

We use as loss function the intersection over unit (IoU) as shown in the following equation:
\begin{equation}
\label{loss}
\mathcal{L}(\Theta)= \frac{1}{N}\sum_{i=0}^{N}\left( 1 - \frac{\hat{\textbf{\textit{X}}}_{seg}^{(i)} \cap \textbf{\textit{X}}_{seg}^{(i)} }{ \hat{\textbf{\textit{X}}}_{seg}^{(i)} \cup \textbf{\textit{X}}_{seg}^{(i)}}\right)
\end{equation}
where $N$ is the number of training samples, $\hat{\textbf{\textit{X}}}_{seg}^{(i)}$ is a prediction mask of the $i$-th input instance and $\textbf{\textit{X}}_{seg}^{(i)}$ is the corresponding ground truth.

As regularization, we adopt dropout\cite{dropout} with a drop rate of $p=0.2$ and early stopping using a 0.1 of the training as validation. We use an Nvidia RTX2080 GPGPU with 8GB of memory, 64GB of DDR4 SDRAM and CUDA 11 with TensorFlow 2.x. With our hardware configuration, the training phase of the segmentation head takes approximately 3 minutes. 

The resulting network is optimized in order to reduce latency, inference cost, memory, and storage
footprint. That is mainly achieved with two distinct techniques: model pruning and quantization. With the first one we simplify the topological structure, removing unnecessary parts of the architecture, or favors a more sparse model introducing zeros to the
parameter tensors. On the other hand, with quantization, we reduce the precision of the numbers used to represent model parameters from float32 to float16. That can be accomplished after the training procedure with a post-training quantization procedure. 

\begin{table}[t]
\caption{Comparison between different devices' energy consumption and inference performances.Graph optimization (G.O.) and weight precision (W.P.) reduction further increase the capability of our already efficient neural network architecture, helping to deal with energy, speed, size, and cost constraints.}
\begin{tabular}{llllll}
\toprule
Device     & GO & WP   & Latency {[}ms{]} & E$_{net}$ {[}mJ{]} & Size {[}MB{]} \\ \hline
RTX 2080   & N  & FP32 & 28 $\pm$ 109        & 819             & 9.3          \\ 
           & Y  & FP32 & 0.1 $\pm$ 0.3    & 52              & 7.4          \\ 
           & Y  & FP16 & 0.1 $\pm$ 0.2      & 39              & 4.9           \\ 
Cortex-A57 & Y  & FP32 & 111 $\pm$ 0.9       & 166             & 4.2           \\
           & Y  & FP16 & 111 $\pm$ 2.3       & 165             & 2.2           \\ 
Cortex-A76 & Y  & FP32 & 55.4 $\pm$ 10.6     & 210             & 4.2           \\ 
           & Y  & FP16 & 65.3 $\pm$ 9.5      & 248             & 2.2           \\ \bottomrule
\end{tabular}
\label{tab:devices_comparison}
\end{table}

In Table \ref{tab:devices_comparison} are summarized experimentation results with some reference architectures. It is possible to notice how weight precision and graph optimization have a high impact on the inference energy consumed by the network and its impact on the main memory of the tested device. 

We evaluate our model using IoU metric with a threshold of 0.9 on the output logits. In this way, we take into account only the values that would be actually used by the SPC algorithm when performing autonomous navigation. On the test set, the final overall metric is 0.62 with a pixel accuracy of 92.7\%. Moreover, considering as true positive only the test instances with a IoU score that exceeds some predefined threshold, we compute the precision values for different IoU thresholds. In particular, precision at 0.4, 0.5, 0.6 and 0.7 are equal to 96.8\%, 85.8\%, 62.8\%, and 24\%, respectively.
Fig. \ref{fig:vegetation_index} allows to qualitative assess the accuracy of the network and its generalization capabilities with three segmentation maps generated from the test site. Moreover, as previously stated, it is possible to exploit the generated segmentation maps not only for navigation but also as a starting point for a vegetative assessment of the crop. Three false-color representations are presented in Fig. \ref{fig:vegetation_index} as examples of vegetation index maps that can be derived directly from reflectances of RGB channels \cite{padua2018multi}. Nevertheless, other spectral bands can be used in conjunction with segmentation maps to extract valuable crop indices.

\subsection{Motion Controller Evaluation}
As already introduced in Section \ref{Dataset_sect}, all collected images are acquired with three fixed orientations of the vineyard rows. That allows us to have more flexibility when extrapolating statistical value to assess the controller performances. More specifically, for each row and for each type of video in the test set, we compute the mean value and standard deviation for the three most meaningful variables of the controller: the abscissa, the linear, and angular velocities. Grouping the calculated values by video class, we obtain results summarized in Table \ref{controller_evaluation}. The outcomes are referred to a frame dimension of 224x224 with maximum linear and angular velocities equal to 1 m/s and 1 rad/s, respectively. Furthermore, we set $l_{depth}= 0.5$, $th_{noise}=0.03 \cdot max(\textbf{\textit{g}}_{noise})$, $\alpha_{EMA}=0.1$, and we fuse $S=3$ segmentation maps before feeding the SPC. In addition, for each video orientation of the test set, we compute the fault rate percentage (FR) of iterations where the control algorithm could not provide any command.

\begin{table}
\caption{Overall motion controller evaluation}
\label{controller_evaluation}
\centering
\begin{tabular}{c c c c c c}
\toprule
Class & Metrics & Abscissa & $v_{x}${[}m/s{]} & $\omega_{z}$ {[}rad/s{]} & FR [\%]\\\hline
\multirow{2}{*}{Center} & $\mu_{center}$    & 111.15   & 0.9926  & 0.0052  & \multirow{2}{*}{0.0}\\ 
                        & $\sigma_{center}$ & 5.05     & 0.0005  & 0.0005  \\\hline
\multirow{2}{*}{Left}   & $\mu_{left}$    & 44.42    & 0.6434  & 0.3566  & \multirow{2}{*}{0.04}\\  
                        & $\sigma_{left}$ & 7.67     & 0.0084  & 0.0083  \\ \hline
\multirow{2}{*}{Right}  & $\mu_{right}$    & 184.93   & 0.5971  & -0.4026 & \multirow{2}{*}{0.26}\\
                        & $\sigma_{right}$ & 12.98    & 0.0114  & 0.0129  \\ \bottomrule
\end{tabular}
\label{default}
\end{table} 

Finally, we run the optimized neural network along with the presented controller on the embedded Jackal's PC. It is mainly composed of a CPU Intel Core i3-4330TE @ $2.4$ GHz and a DDR3 RAM of 4GB. In such conditions, the Intel Realsense D435i provides both RBG images and depth map at $30$ FPS, while the optimized deep neural network is able to process the RGB images on the CPU at $22$ FPS, and the controller generates velocity commands at $5$ Hz. All considered, the overall performances of the deployed algorithm are very promising. Indeed, five output commands per second are greatly sufficient taking into account the slow dynamic of the vehicle.

\subsection{Simulation Environment Results}
\begin{figure}
    \centering
    \subfloat[]
    {
        \includegraphics[scale=0.17]{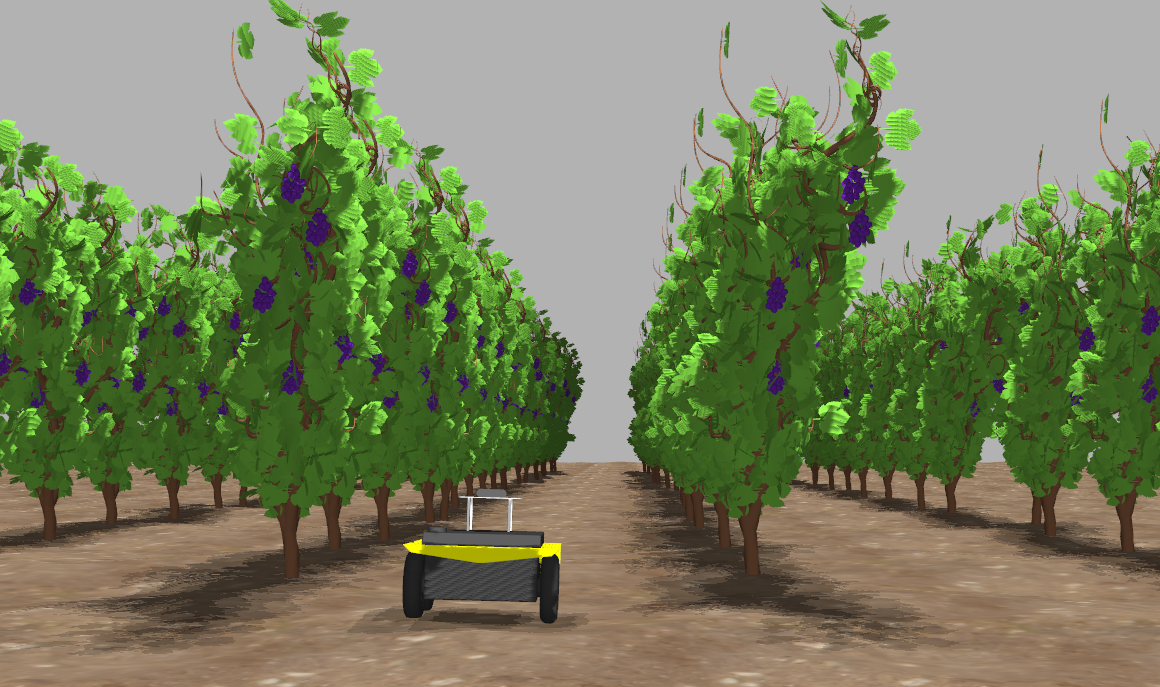}
        \label{fig:vine_top}
    }\\
    \subfloat[]
    {
        \includegraphics[scale=0.06]{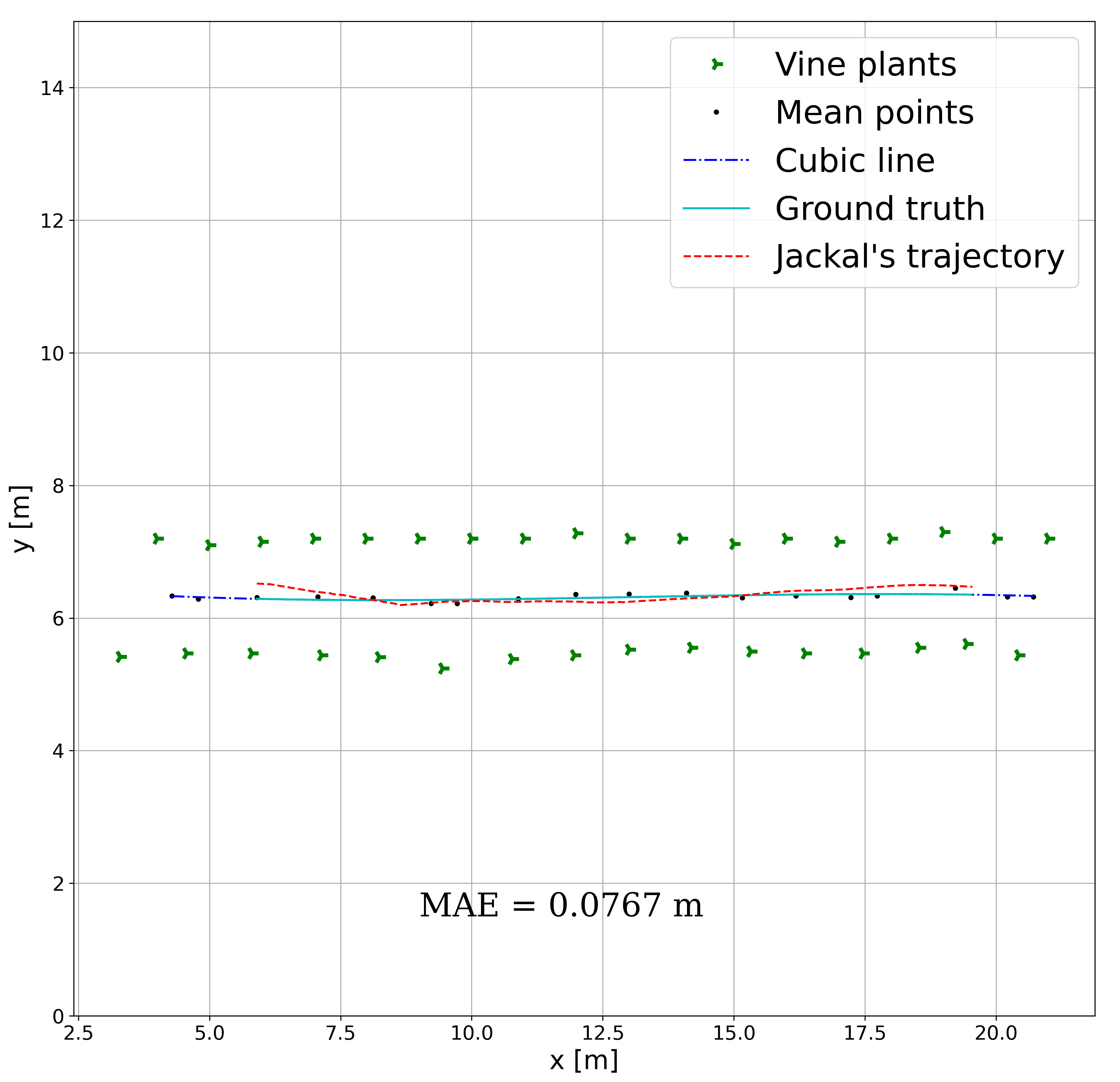}
        \label{fig:first_result}
    }
    \caption{(\textbf{a}) shows the straight vine rows of the first simulation environment. While (\textbf{b}) summarizes the results coming from the first simulation environment. The most important aspects are: the ground truth (cyan line) and the Jackal's pose over time (dash red line). The black points represent the mid-distance between vine rows, while the blue line is an approximation of such points.}
    \label{fig:first_env}
\end{figure}
\begin{figure}
\centering
    \subfloat[]
    {
        \includegraphics[scale=0.14]{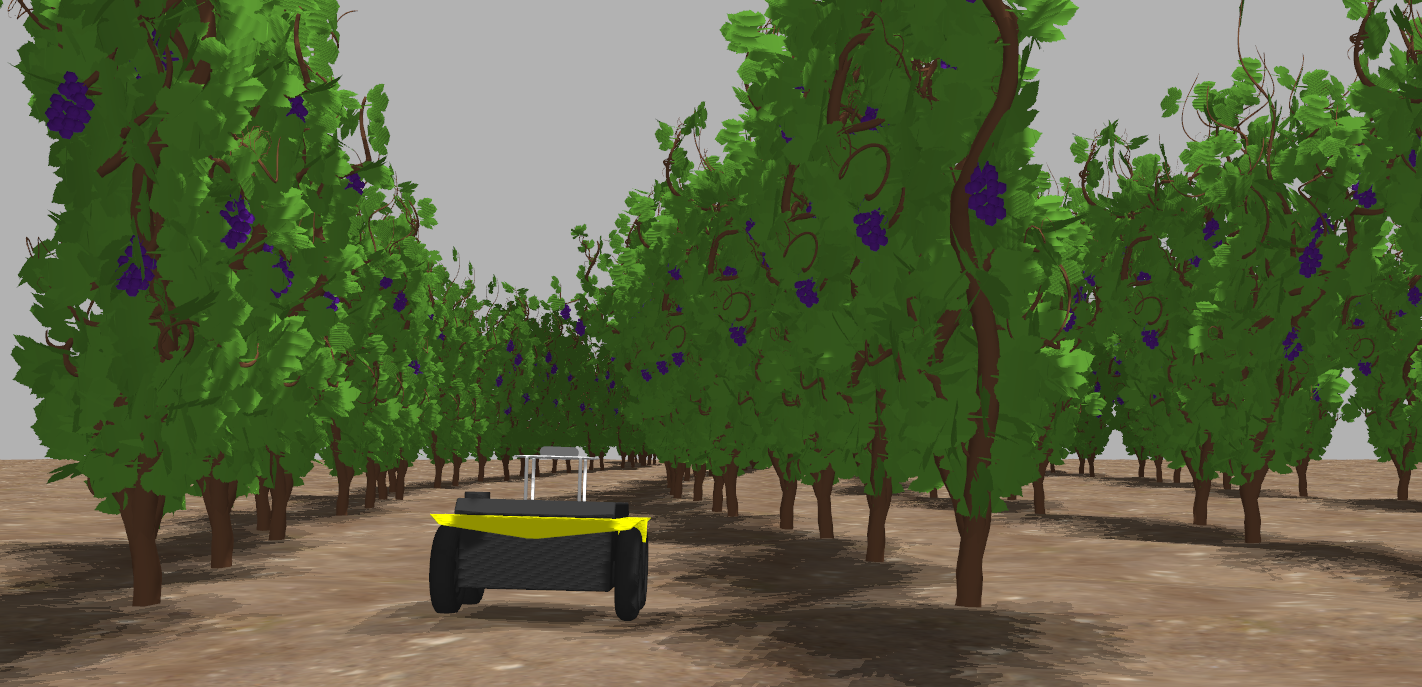}
        \label{fig:curved_vine_top}
    }\\
    \subfloat[]
    {
        \includegraphics[scale=0.06]{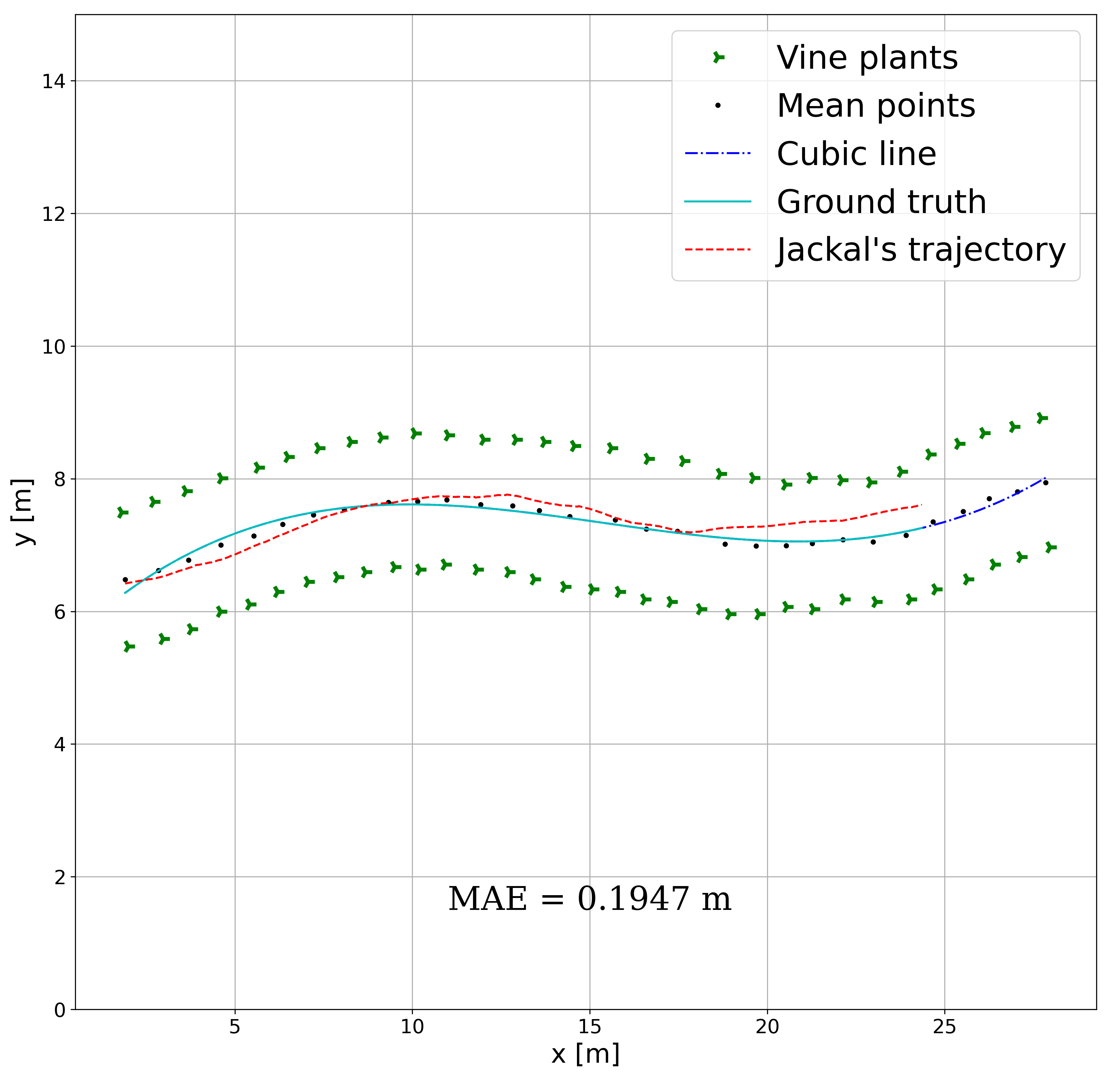}
        \label{fig:second_result}
    }
    \caption{(\textbf{a}) shows the slight curvature of vine rows in the second simulated environment. (\textbf{b}) contains the relevant results obtained from the second simulation environment. The black points represent the mid-distance between vine rows, while the blue line is an approximation of such points. Moreover, this chart allows to visualize the comparison between the ground truth (cyan line) and the UGV's trajectory (dash red line).}
    \label{fig:second_env}
\end{figure}

In addition to previous results, the controller algorithm is tested in two custom simulated environments. As a simulator engine, we use Gazebo\footnote{http://gazebosim.org/}, which is ROS-compatible and allows us to customize the simulation environments and exploit several plugins to simulate sensors(e.g. cameras). The simulation model of the Jackal UGV is provided by Clearpath Robotics through URDF file, which contains specification about mechanical structure and links related to the robot platform. On the other hand, the simulation environments are designed from scratch; drawing a completely custom vine plant and arranging the vine rows on a plane in order to realistically reproduce the vineyard geometry, as shown in Fig. \ref{fig:vine_top} and Fig. \ref{fig:curved_vine_top}. The inter-row distance ranges from 1.70 meters to 2.00 meters, while the vine plants' distance in the same row ranges from 0.70 meters to 1.0 meters from each other. The simulated plane reproduces the bumpy and uneven terrain of a real vineyard exploiting the heightmap option of the SDF format for model building in Gazebo. The involved simulated camera is an Intel Realsense d435i, that by means a Gazebo plugin is able to provide RGB frames and depth map of what the UGV is seeing. It is placed 10 centimeters up and with 0 degrees of tilt with respect to Jackal's plate.\\
The real trajectory followed by the Jackal has been obtained using a Gazebo plugin (\textit{libgazebo\_ros\_p3d}), which provides the real pose of the robot inside the simulation environment so that it is possible to perform additional analysis on the followed path. The procedure to measure the ability of the proposed controller to maintain the vine rows centrality is very straightforward. First, the midline between two vine rows is obtained averaging the Euclidean distances of the points belonging to a vine row with respect to the nearest point related to the other vine row, and approximating such distances with a cubic line.
Subsequently, taking the midline as ground truth and the Jackal's trajectory, the Mean Absolute Error (MAE) is computed to measure the quality of the path.

The first simulation environment consists of straight vine rows, where the proposed controller has achieved excellent results obtaining a MAE$=0.0767$ m over several trials. While, the second simulated vineyard is composed of vine rows with curvatures, where the controller slightly worsens the performances achieving a MAE of $0.1947$ m. Nevertheless, results are still very promising, taking into account the challenging environment.
   
\section{CONCLUSIONS}

We introduced a novel low-cost motion controller algorithm driven by edge AI image segmentation that showed very promising results in motion control robustness and segmentation accuracy, solely relying on low-range equipment. It can be easily deployed on a variety of mobile platforms thanks to the low computational load required by the overall processing framework. Finally, generated semantic information of the navigation environment can be directly exploited for a vegetative assessment of the crop. Further works will aim at assessing our proposed module inside a broader navigation framework devised by our research group for affordable autonomous vineyard navigation.





\section*{ACKNOWLEDGMENT}
This work has been developed with the contribution of the Politecnico di Torino Interdepartmental Centre for Service Robotics PIC4SeR\footnote{https://pic4ser.polito.it} and SmartData@Polito\footnote{https://smartdata.polito.it}.


\bibliography{IEEEabrv,mybib}

\begin{thebibliography}{10}
\providecommand{\url}[1]{#1}
\csname url@rmstyle\endcsname
\providecommand{\newblock}{\relax}
\providecommand{\bibinfo}[2]{#2}
\providecommand\BIBentrySTDinterwordspacing{\spaceskip=0pt\relax}
\providecommand\BIBentryALTinterwordstretchfactor{4}
\providecommand\BIBentryALTinterwordspacing{\spaceskip=\fontdimen2\font plus
\BIBentryALTinterwordstretchfactor\fontdimen3\font minus
  \fontdimen4\font\relax}
\providecommand\BIBforeignlanguage[2]{{%
\expandafter\ifx\csname l@#1\endcsname\relax
\typeout{** WARNING: IEEEtran.bst: No hyphenation pattern has been}%
\typeout{** loaded for the language `#1'. Using the pattern for}%
\typeout{** the default language instead.}%
\else
\language=\csname l@#1\endcsname
\fi
#2}}

\bibitem{harvesting}
C.~W. Bac, E.~J. van Henten, J.~Hemming, and Y.~Edan, ``Harvesting robots for
  high-value crops: State-of-the-art review and challenges ahead,''
  \emph{Journal of Field Robotics}, vol.~31, no.~6, pp. 888--911, 2014.

\bibitem{spraygrape}
R.~Berenstein, O.~B. Shahar, A.~Shapiro, and Y.~Edan, ``Grape clusters and
  foliage detection algorithms for autonomous selective vineyard sprayer,''
  \emph{Intelligent Service Robotics}, vol.~3, no.~4, pp. 233--243, 2010.

\bibitem{seeding}
J.~Katupitiya, R.~Eaton, and T.~Yaqub, ``Systems engineering approach to
  agricultural automation: new developments,'' in \emph{2007 1st Annual IEEE
  Systems Conference}.\hskip 1em plus 0.5em minus 0.4em\relax IEEE, 2007, pp.
  1--7.

\bibitem{irrigation}
D.~Kohanbash, A.~Valada, and G.~Kantor, ``Irrigation control methods for
  wireless sensor network,'' in \emph{2012 Dallas, Texas, July 29-August 1,
  2012}.\hskip 1em plus 0.5em minus 0.4em\relax American Society of
  Agricultural and Biological Engineers, 2012, p.~1.

\bibitem{monitoring}
N.~Virlet, K.~Sabermanesh, P.~Sadeghi-Tehran, and M.~J. Hawkesford, ``Field
  scanalyzer: an automated robotic field phenotyping platform for detailed crop
  monitoring,'' \emph{Functional Plant Biology}, vol.~44, no.~1, pp. 143--153,
  2017.

\bibitem{lidar_gps}
F.~Callegati, A.~Samor{\`i}, R.~Tazzari, N.~Mimmo, and L.~Marconi, ``Autonomous
  tracked agricultural ugv configuration and navigation experimental results,''
  2018.

\bibitem{lidar_based_nav}
\BIBentryALTinterwordspacing
P.~M. Blok, K.~{van Boheemen}, F.~K. {van Evert}, J.~IJsselmuiden, and G.-H.
  Kim, ``Robot navigation in orchards with localization based on particle
  filter and kalman filter,'' \emph{Computers and Electronics in Agriculture},
  vol. 157, pp. 261--269, 2019. [Online]. Available:
  \url{https://www.sciencedirect.com/science/article/pii/S0168169918315230}
\BIBentrySTDinterwordspacing

\bibitem{lidar_based_nav2}
\BIBentryALTinterwordspacing
F.~B. Malavazi, R.~Guyonneau, J.-B. Fasquel, S.~Lagrange, and F.~Mercier,
  ``Lidar-only based navigation algorithm for an autonomous agricultural
  robot,'' \emph{Computers and Electronics in Agriculture}, vol. 154, pp.
  71--79, 2018. [Online]. Available:
  \url{https://www.sciencedirect.com/science/article/pii/S0168169918302679}
\BIBentrySTDinterwordspacing

\bibitem{RADCLIFFE2018165}
J.~Radcliffe, J.~Cox, and D.~M. Bulanon, ``Machine vision for orchard
  navigation,'' \emph{Computers in Industry}, vol.~98, pp. 165--171, 2018.

\bibitem{aghi2020autonomous}
D.~Aghi, V.~Mazzia, and M.~Chiaberge, ``Autonomous navigation in vineyards with
  deep learning at the edge,'' in \emph{International Conference on Robotics in
  Alpe-Adria Danube Region}.\hskip 1em plus 0.5em minus 0.4em\relax Springer,
  2020, pp. 479--486.

\bibitem{gps_only}
O.~{Ly}, H.~{Gimbert}, G.~{Passault}, and G.~{Baron}, ``A fully autonomous
  robot for putting posts for trellising vineyard with centimetric accuracy,''
  in \emph{2015 IEEE International Conference on Autonomous Robot Systems and
  Competitions}, 2015, pp. 44--49.

\bibitem{lidar_gps2}
S.~J. Moorehead, C.~K. Wellington, B.~J. Gilmore, and C.~Vallespi, ``Automating
  orchards: A system of autonomous tractors for orchard maintenance,'' in
  \emph{Proceedings of the IEEE International Conference of Intelligent Robots
  and Systems, Workshop on Agricultural Robotics}, 2012.

\bibitem{hansen2011orchard}
S.~Hansen, E.~Bayramoglu, J.~C. Andersen, O.~Ravn, N.~Andersen, and N.~K.
  Poulsen, ``Orchard navigation using derivative free kalman filtering,'' in
  \emph{Proceedings of the 2011 American Control Conference}.\hskip 1em plus
  0.5em minus 0.4em\relax IEEE, 2011, pp. 4679--4684.

\bibitem{GPS_accuracy}
\BIBentryALTinterwordspacing
M.~S.~N. Kabir, M.-Z. Song, N.-S. Sung, S.-O. Chung, Y.-J. Kim, N.~Noguchi, and
  S.-J. Hong, ``Performance comparison of single and multi-gnss receivers under
  agricultural fields in korea,'' \emph{Engineering in Agriculture, Environment
  and Food}, vol.~9, no.~1, pp. 27--35, 2016. [Online]. Available:
  \url{https://www.sciencedirect.com/science/article/pii/S1881836615300136}
\BIBentrySTDinterwordspacing

\bibitem{gpsunreliable}
S.~Marden and M.~Whitty, ``Gps-free localisation and navigation of an unmanned
  ground vehicle for yield forecasting in a vineyard,'' in \emph{Recent
  Advances in Agricultural Robotics, International workshop collocated with the
  13th International Conference on Intelligent Autonomous Systems (IAS-13)},
  2014.

\bibitem{echord}
P.~Astolfi, A.~Gabrielli, L.~Bascetta, and M.~Matteucci, ``Vineyard autonomous
  navigation in the echord++ grape experiment,'' \emph{IFAC-PapersOnLine},
  vol.~51, no.~11, pp. 704--709, 2018.

\bibitem{zaman2019cost}
S.~Zaman, L.~Comba, A.~Biglia, D.~R. Aimonino, P.~Barge, and P.~Gay,
  ``Cost-effective visual odometry system for vehicle motion control in
  agricultural environments,'' \emph{Computers and Electronics in Agriculture},
  vol. 162, pp. 82--94, 2019.

\bibitem{KAMILARIS201870}
\BIBentryALTinterwordspacing
A.~Kamilaris and F.~X. Prenafeta-Boldú, ``Deep learning in agriculture: A
  survey,'' \emph{Computers and Electronics in Agriculture}, vol. 147, pp.
  70--90, 2018. [Online]. Available:
  \url{https://www.sciencedirect.com/science/article/pii/S0168169917308803}
\BIBentrySTDinterwordspacing

\bibitem{cropclass1}
N.~Kussul, M.~Lavreniuk, S.~Skakun, and A.~Shelestov, ``Deep learning
  classification of land cover and crop types using remote sensing data,''
  \emph{IEEE Geoscience and Remote Sensing Letters}, vol.~14, no.~5, pp.
  778--782, 2017.

\bibitem{cropclass2}
A.~K. Mortensen, M.~Dyrmann, H.~Karstoft, R.~N. J{\o}rgensen, R.~Gislum,
  \emph{et~al.}, ``Semantic segmentation of mixed crops using deep
  convolutional neural network.'' in \emph{CIGR-AgEng Conference, 26-29 June
  2016, Aarhus, Denmark. Abstracts and Full papers}.\hskip 1em plus 0.5em minus
  0.4em\relax Organising Committee, CIGR 2016, 2016, pp. 1--6.

\bibitem{mazzia2020improvement}
V.~Mazzia, A.~Khaliq, and M.~Chiaberge, ``Improvement in land cover and crop
  classification based on temporal features learning from sentinel-2 data using
  recurrent-convolutional neural network (r-cnn),'' \emph{Applied Sciences},
  vol.~10, no.~1, p. 238, 2020.

\bibitem{yield_estimation}
Q.~Wang, S.~Nuske, M.~Bergerman, and S.~Singh, ``Automated crop yield
  estimation for apple orchards,'' in \emph{Experimental robotics}.\hskip 1em
  plus 0.5em minus 0.4em\relax Springer, 2013, pp. 745--758.

\bibitem{estimators2}
K.~Kuwata and R.~Shibasaki, ``Estimating crop yields with deep learning and
  remotely sensed data,'' in \emph{2015 IEEE International Geoscience and
  Remote Sensing Symposium (IGARSS)}.\hskip 1em plus 0.5em minus 0.4em\relax
  IEEE, 2015, pp. 858--861.

\bibitem{mazzia2020uav}
V.~Mazzia, L.~Comba, A.~Khaliq, M.~Chiaberge, and P.~Gay, ``Uav and machine
  learning based refinement of a satellite-driven vegetation index for
  precision agriculture,'' \emph{Sensors}, vol.~20, no.~9, p. 2530, 2020.

\bibitem{desease_detector}
R.~Calder{\'o}n, J.~A. Navas-Cort{\'e}s, and P.~J. Zarco-Tejada, ``Early
  detection and quantification of verticillium wilt in olive using
  hyperspectral and thermal imagery over large areas,'' \emph{Remote Sensing},
  vol.~7, no.~5, pp. 5584--5610, 2015.

\bibitem{leaf_desease}
M.~L{\'o}pez-L{\'o}pez, R.~Calder{\'o}n, V.~Gonz{\'a}lez-Dugo, P.~J.
  Zarco-Tejada, and E.~Fereres, ``Early detection and quantification of almond
  red leaf blotch using high-resolution hyperspectral and thermal imagery,''
  \emph{Remote Sensing}, vol.~8, no.~4, p. 276, 2016.

\bibitem{fruit_desease}
Y.~Tian, G.~Yang, Z.~Wang, E.~Li, and Z.~Liang, ``Detection of apple lesions in
  orchards based on deep learning methods of cyclegan and yolov3-dense,''
  \emph{Journal of Sensors}, vol. 2019, 2019.

\bibitem{deepway}
V.~Mazzia, F.~Salvetti, D.~Aghi, and M.~Chiaberge, ``Deepway: a deep learning
  waypoint estimator for global path generation,'' 2021.

\bibitem{simo_thesis}
S.~Cerrato, ``Gps-based autonomous navigation of unmanned ground vehicles in
  precision agriculture applications,'' Master's thesis, Politecnico di Torino,
  2020.

\bibitem{mobnet3}
A.~Howard, M.~Sandler, G.~Chu, L.-C. Chen, B.~Chen, M.~Tan, W.~Wang, Y.~Zhu,
  R.~Pang, V.~Vasudevan, \emph{et~al.}, ``Searching for mobilenetv3,'' in
  \emph{Proceedings of the IEEE/CVF International Conference on Computer
  Vision}, 2019, pp. 1314--1324.

\bibitem{mobilenetv2}
M.~Sandler, A.~Howard, M.~Zhu, A.~Zhmoginov, and L.-C. Chen, ``Mobilenetv2:
  Inverted residuals and linear bottlenecks,'' in \emph{Proceedings of the IEEE
  conference on computer vision and pattern recognition}, 2018, pp. 4510--4520.

\bibitem{squeeze}
J.~Hu, L.~Shen, and G.~Sun, ``Squeeze-and-excitation networks,'' in
  \emph{Proceedings of the IEEE conference on computer vision and pattern
  recognition}, 2018, pp. 7132--7141.

\bibitem{chen2014semantic}
L.-C. Chen, G.~Papandreou, I.~Kokkinos, K.~Murphy, and A.~L. Yuille, ``Semantic
  image segmentation with deep convolutional nets and fully connected crfs,''
  \emph{arXiv preprint arXiv:1412.7062}, 2014.

\bibitem{chen2017rethinking}
L.-C. Chen, G.~Papandreou, F.~Schroff, and H.~Adam, ``Rethinking atrous
  convolution for semantic image segmentation,'' \emph{arXiv preprint
  arXiv:1706.05587}, 2017.

\bibitem{chen2017deeplab}
L.-C. Chen, G.~Papandreou, I.~Kokkinos, K.~Murphy, and A.~L. Yuille, ``Deeplab:
  Semantic image segmentation with deep convolutional nets, atrous convolution,
  and fully connected crfs,'' \emph{IEEE transactions on pattern analysis and
  machine intelligence}, vol.~40, no.~4, pp. 834--848, 2017.

\bibitem{skip}
J.~Long, E.~Shelhamer, and T.~Darrell, ``Fully convolutional networks for
  semantic segmentation,'' in \emph{Proceedings of the IEEE Conference on
  Computer Vision and Pattern Recognition (CVPR)}, June 2015.

\bibitem{aghi2020local}
D.~Aghi, V.~Mazzia, and M.~Chiaberge, ``Local motion planner for autonomous
  navigation in vineyards with a rgb-d camera-based algorithm and deep learning
  synergy,'' \emph{Machines}, vol.~8, no.~2, p.~27, 2020.

\bibitem{dutta2019vgg}
\BIBentryALTinterwordspacing
A.~Dutta and A.~Zisserman, ``The {VIA} annotation software for images, audio
  and video,'' in \emph{Proceedings of the 27th ACM International Conference on
  Multimedia}, ser. MM '19.\hskip 1em plus 0.5em minus 0.4em\relax New York,
  NY, USA: ACM, 2019. [Online]. Available:
  \url{https://doi.org/10.1145/3343031.3350535}
\BIBentrySTDinterwordspacing

\bibitem{gaussian}
D.~A. Reynolds, ``Gaussian mixture models.'' \emph{Encyclopedia of biometrics},
  vol. 741, pp. 659--663, 2009.

\bibitem{padua2018multi}
L.~P{\'a}dua, P.~Marques, J.~Hru{\v{s}}ka, T.~Ad{\~a}o, E.~Peres, R.~Morais,
  and J.~J. Sousa, ``Multi-temporal vineyard monitoring through uav-based rgb
  imagery,'' \emph{Remote Sensing}, vol.~10, no.~12, p. 1907, 2018.

\bibitem{tl}
K.~Weiss, T.~M. Khoshgoftaar, and D.~Wang, ``A survey of transfer learning,''
  \emph{Journal of Big data}, vol.~3, no.~1, pp. 1--40, 2016.

\bibitem{russakovsky2015imagenet}
O.~Russakovsky, J.~Deng, H.~Su, J.~Krause, S.~Satheesh, S.~Ma, Z.~Huang,
  A.~Karpathy, A.~Khosla, M.~Bernstein, \emph{et~al.}, ``Imagenet large scale
  visual recognition challenge,'' \emph{International journal of computer
  vision}, vol. 115, no.~3, pp. 211--252, 2015.

\bibitem{dropout}
N.~Srivastava, G.~Hinton, A.~Krizhevsky, I.~Sutskever, and R.~Salakhutdinov,
  ``Dropout: a simple way to prevent neural networks from overfitting,''
  \emph{The journal of machine learning research}, vol.~15, no.~1, pp.
  1929--1958, 2014.

\end{thebibliography}
\bibliographystyle{IEEEtran}


\end{document}